\documentclass[journal]{IEEEtran}

\usepackage{cite}

\ifCLASSINFOpdf
\else
\fi

\usepackage{times}
\usepackage{epsfig}
\usepackage{graphicx}
\usepackage{amssymb}
\usepackage{bm}
\usepackage{bbm}
\usepackage{multirow}
\usepackage{bbding}
\usepackage{pifont}
\usepackage{booktabs}
\usepackage{color}
\usepackage{subfig}

\usepackage[normalem]{ulem}

\usepackage{amsmath}
\newcommand{\cmark}{}%
\newcommand{\xmark}{\ding{55}}%

\begin{document}
\title{Learning to Update for Object Tracking with Recurrent Meta-learner}

\author{Bi~Li,
		Wenxuan~Xie,
		Wenjun~Zeng,~\IEEEmembership{Fellow,~IEEE,}
        and~Wenyu~Liu,~\IEEEmembership{Senior Member,~IEEE}%
\thanks{B.~Li and W.~Liu are with the School of Electronic Information and
Communications, Huazhong University of Science and Technology,
Wuhan 430074, China. (e-mail: libi@hust.edu.cn; liuwy@hust.edu.cn)}%
\thanks{W.~Xie and W.~Zeng are with the Microsoft Research Asia, Beijing 100080,
China.(e-mail: wenxie@microsoft.com; wezeng@microsoft.com)}%
}

\maketitle

\begin{abstract}
  Model update lies at the heart of object tracking.
  Generally, model update is formulated as an online learning problem where a target model is learned over the online training set. Our key innovation is to \emph{formulate the model update problem in the meta-learning framework and learn the online learning algorithm itself using large numbers of offline videos}, i.e., \emph{learning to update}.
 The learned updater takes as input the online training set and outputs an updated target model. 
  As a first attempt, we design the learned updater based on recurrent neural networks (RNNs) and demonstrate its application in a template-based tracker and a correlation filter-based tracker. Our learned updater consistently improves the base trackers and runs faster than realtime on GPU while requiring small memory footprint during testing.
  Experiments on standard benchmarks demonstrate that our learned updater outperforms commonly used update baselines including the efficient exponential moving average (EMA)-based update and the well-designed stochastic gradient descent (SGD)-based update. Equipped with our learned updater, the template-based tracker achieves state-of-the-art performance among realtime trackers on GPU.
\end{abstract}


\IEEEpeerreviewmaketitle

\section{Introduction}
\IEEEPARstart{O}{bject} tracking is a crucial task in computer vision that deals with the problem of localizing one arbitrary target object in a video, given only the target position in the first frame.  Typically, bounding boxes are used for representing the target position. Arbitrary target object implies that a dedicated target model\footnote{In this work, we only consider scoring-based target model that outputs the confidence of an image patch being the target. Note there are also position-regression based target models that, given an image patch, directly regress the target position. } is needed for each target during testing. This is typically accomplished through online learning where the online training set\footnote{The framework of learning to update contains an offline phase and an online phase. The offline phase is referred to as training the updater with offline videos, whereas the online phase refers to tracking a new sequence. The term ``(online) training set'' means the set of image patches collected for updating the target model (during the online phase).} is extracted from the test video and a target model is initialized and updated on the fly. 

In this work, we tackle the problem of model update: after an initial target model is built using reliable supervision in the first frame, how to exploit information in subsequent frames and update the initial model along with tracking? Model update is challenging because of \emph{unreliable and highly correlated online training set}. The only reliable supervision for building a target model is the information in the first frame. After that, the online training set is collected based on predicted target position, which is not always reliable. When small errors in the training samples accumulate, model update can cause the drifting problem. Moreover, the extracted online training set is highly correlated since most of the training samples are simply the translated and scaled version of a base sample. Correlated samples are easy to fit and do not help much with the generalization to hard samples.

Recent works \cite{tao2016siamese, bertinetto2016fully} have investigated the possibility of no model update at all, and achieved remarkable tracking performance. These approaches can be interpreted as learning an invariant and discriminative feature extractor such that the target remains stable in the feature space and is separable from the background. However, learning a representation that is both invariant and discriminative for a long time is intrinsically difficult, as with time evolves, features that once are discriminative may become irrelevant and vice versa. Consider that when a red car drives into a dark tunnel, the red color becomes irrelevant, although it is discriminative before entering the tunnel. Instead of striving to construct a perfect model at the first frame, model update tries to keep up with the current target appearance along with tracking by constantly incorporating the new target information, and therefore eases the burden of feature representation. Moreover, by gradually adapting to the current video context, the tracking problem can be considerably simplified \cite{grabner2008semi}. In scenarios where target exhibits multi-modality, model update is indispensable.

Generally, model update can be formulated as an online learning problem with two stages. First, an online training set is collected along with tracking. Then, the target model is learned on the training set using algorithms like stochastic gradient descent (SGD). Existing update methods typically suffer from the problem of large training set and slow convergence thus being too slow for practical use. Moreover, due to unreliable training data, regularizations and rules are carefully designed based on expertise in the field to avoid model drifting.

\begin{figure}
	\centering
	\includegraphics[width=1\columnwidth]{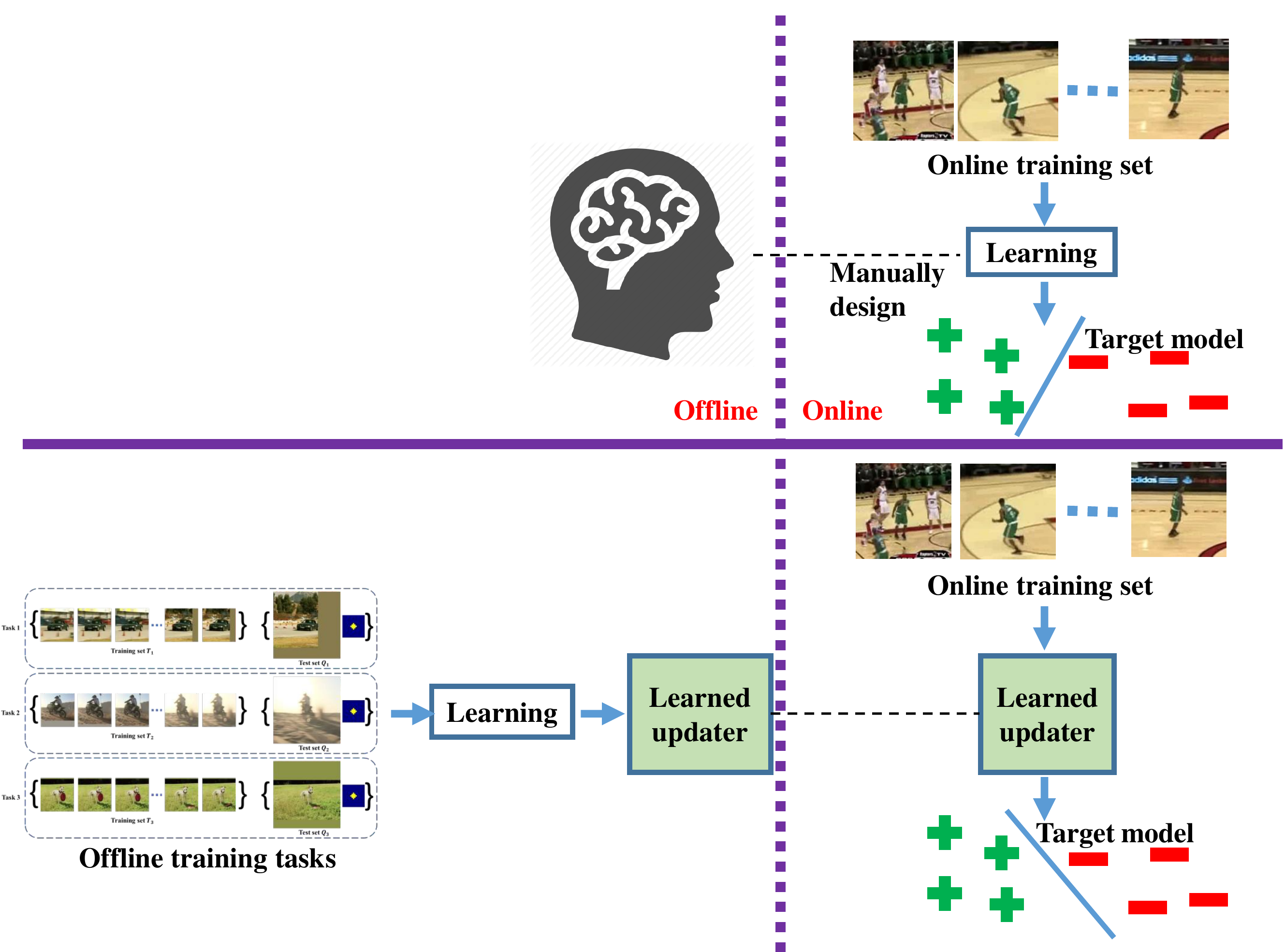}
	\caption{A general introduction of learning to update. \textbf{Top}: Typically, a learning algorithm (e.g., SGD) is manually designed for online learning of the target model. \textbf{Bottom}: Contrarily, a learned updater is adopted for online learning, which is offline-trained using large numbers of videos. All figures are best viewed in color.}
	\label{fig:learn_to_update}
\end{figure}

In this work, we advocate the paradigm for object tracking that \emph{eases the heavy burdens of online learning by offline learning}. Offline learning is performed before the actual tracking takes place and the learned model is shared among all test videos. Online learning, in contrast, is conducted during tracking and the learned model is specific to each test video. Our key innovation is to \emph{formulate the model update problem in the meta-learning framework and} learn the online learning algorithm itself using large numbers of offline videos, i.e., \emph{learning to update}. The offline-learned update method, which we call the learned updater, takes in the online training set and outputs the updated target model. Please refer to Fig.~\ref{fig:learn_to_update} for visual illustrations.

The benefit of learning to update is threefold: 1) After seeing all kinds of target variations in the offline training phase, the learned updater is able to capture \emph{target variation patterns} among videos. These learned patterns are implicitly used during testing to avoid unlikely update (e.g., update to background) and thus can be seen as a form of regularization, which enables our learned updater to handle the unreliable online training set. 2) The learned updater is able to update the target model based on not only the online training set, but also rules learned from the offline dataset. Therefore, the learned updater is able to see beyond the highly correlated online training set and makes the updated model capable of generalizing to more challenging scenarios. 3) The learned patterns enable fast inference of the learned updater. As a result, our learned updater improves the performance of base trackers while running faster than realtime on GPU with a single forward pass of the neural network per frame. 

In this paper, we formulate model update as a meta-learning problem (a.k.a learning to learn) \cite{Andrychowicz2016LearningTL, ravi2016optimization} and learn a model updater. Specifically, our learned updater is embodied as a RNN, which is well known for its ability to model sequential/temporal variations. Previous efforts to model target variations based on RNNs mostly fail to deliver satisfactory tracking performance due to inadequate offline training videos. In this work, we contribute several techniques to overcome data deficiencies and train RNNs effectively. With a properly trained updater, our tracker achieves state-of-the-art performance among realtime trackers. 

As a first attempt of learning to update for object tracking, we demonstrate its application on two base trackers: a template-based tracker for its simplicity and a correlation filter-based tracker for its wide adoption. Our learned updater considerably improves the base trackers and outperforms relevant model update baselines including the exponential moving average (EMA)- and the SGD-based update method. 

In summary, our contributions are threefold: 1) We propose a novel model update method for object tracking that (i) is formulated as a meta-learning problem, capable of learning target variation patterns and facilitating effective tracking, and (ii) runs faster than realtime (82 fps with SiamFC tracker and 70 fps with CFNet tracker) while requiring small memory footprint, thus being suitable for practical applications;  2) We propose several techniques to train our RNN-based updater effectively; 3) We validate our method in common object tracking benchmarks and show that it (i) consistently outperforms relevant model update baselines, and (ii) obtains state-of-the-art performance among realtime trackers.

\section{Related Work}
\textbf{Handcrafted Update Methods}.
In general, target variation can be decomposed as short- and long-term variation. \cite{Jepson2001RobustOA} proposes a probabilistic mixture model, which has a stable component to account for the long-term variation, a wandering component and a loss component for the short-term variation. Inspired by the Atkinson-Shiffrin Memory Model, \cite{Hong2015MUltiStoreT} uses short- and long-term memory to handle target variations. Correlation filter and keypoint matching are employed for short- and long-term memory, respectively. Instead of using two separate components, we design a single component based on RNN and learn to process short- and long-term information in a data-driven manner.  

One critical problem in model update is related to the stability-plasticity dilemma \cite{Grossberg1987CompetitiveLF}. On one hand, model update should be stable to avoid the drifting problem where small errors accumulate and the model gets adapted to other objects. On the other hand, it also needs plasticity to effectively assimilate new information derived during tracking. \cite{matthews2004template} coined it the \emph{template update problem}. They adopt a conservative update strategy which keeps the target model in the first frame (i.e., initial model), and updates the latest model only if its predicted locations are close to those of the initial model. Similar techniques are adopted in \cite{guo2017learning} which learns a ridge regression based on the first and the last target model. Inspired by this, we design an anchor loss that uses the first target model as an anchor point. We find it particularly useful in our learning based updater and will elaborate on it in Section~\ref{sec:learning-to-update}. 

Another strategy to handle the model drifting problem is being more careful about the derived training samples. Instead of making hard decisions about labeling training samples as target or background, \cite{grabner2008semi} proposes a semi-supervised approach where only samples from the first frame are labeled and training samples from subsequent frames remain unlabeled. \cite{kwak2011learning} proposes an occlusion detector and updates the model only if the occlusion level is low. \cite{Babenko2011RobustOT} uses multiple-instance learning to update model with bags of samples, where positive bag contains at least one positive sample (without knowing which one), and negative bag contains only negative samples. \cite{Danelljan2016AdaptiveDO} estimates the training sample qualities by optimizing the target model loss with respect to both the target model and the sample qualities. We facilitate training information selection by adopting a gating mechanism in our learned updater.

A popular model update method for correlation filter-based trackers \cite{bolme2010visual, henriques2015high} is exponential moving average (EMA), which performs linear interpolation from the newly trained model (using only training samples in the current frame) and the previous target model. This method is attractive because 1) the update process is highly efficient without iterative optimization and 2) training samples are processed on the fly without the need to be stored. However, it is unlikely that linear combination can capture all of the complex target variations. Our model update method outperforms EMA-based update while preserving all the practical benefits mentioned above.

\textbf{Meta-learning}.
In this work, model update is formulated as a meta-learning problem. Essentially, meta-learning models a learning problem in two scales: learning for specific tasks, and learning for general patterns that rule specific tasks. In our case, updating target model for a specific target (e.g., an airplane) is a specific task. We aim to learn a model update method that is applicable to any specific tasks. \cite{Andrychowicz2016LearningTL, ravi2016optimization} learn to solve the optimization problem of neural networks based on RNNs. Their methods mainly focus on fast convergence and are not readily applicable for model update due to the unreliable training samples. 

\textbf{RNN-based trackers}.
RNNs are well known for their ability to model temporal variations. Given the importance of modeling temporal variations of target in object tracking, it is natural to consider taking advantage of RNNs. \cite{gan2015first} is among the first to use RNN for object tracking, but has only shown to work on simple synthetic datasets. RATM \cite{kahou2015ratm} and HART \cite{Kosiorek2017HierarchicalAR} develop attention mechanisms based on RNNs and demonstrate success on natural image datasets KTH \cite{Schldt2004RecognizingHA} and KITTI \cite{Geiger2013VisionMR}. Re3 \cite{gordon2017re3} models both appearance and motion variations using RNNs and achieves comparable results on several object tracking benchmarks \cite{Kristan2014TheVO, Kristan2016TheVO, wu2013online}. However, Re3 \cite{gordon2017re3} only models short-term variations and requires manual resetting of RNN states every 32 frames. RFL \cite{yang2017recurrent} proposes a filter generation method based on RNNs and resembles our method applied to the template-based tracker. Nevertheless, we tackle a more general model update problem and formulate our method in the meta-learning framework. RFL fails to deliver satisfactory tracking performance due to aggressive updating which is common in RNN-based trackers. By formulating the model update in the meta-learning framework, we focus on the generalization ability of the online-learned target model. With the proposed anchor loss, our approach outperforms RFL by a large margin. To the best of our knowledge, we propose the first RNN-based tracker that achieves state-of-the-art tracking performance among GPU-based realtime trackers.

\section{Base Trackers and Base Update Methods} \label{sec:base-tracker-base-update-method}
In this work, we focus on the update of \emph{linear target model} due to its simplicity and wide adoption.  

Basically, a target model is a scoring function that outputs the confidence of an input image patch being the target. Importantly, the target model should have parameters $\theta$ that is \emph{updatable}. By model update, we mean updating $\theta$ such that it accommodates the target variations during tracking.
 
For a linear target model, the confidence $c$ of an image patch with feature $x$ is the inner product between $\theta$ and $x$, i.e., $c = \theta^T x$.

During tracking, a tracker gathers a training set $T$ for updating the target model parameters $\theta$. Since the dataset is gathered online, it is called the online training set. Let $T_t$ be the online training set at time $t$, the target model is updated by an update function $u$, i.e., $\theta_t = u(T_t)$. This work is about \emph{learning} an update function $u_\phi(\cdot)$ with parameters $\phi$ using large numbers of offline videos.

Before diving into the proposed learned updater, we introduce two base trackers with linear target model as well as two baseline model update methods in this section. 

\subsection{Template-Based Tracker: SiamFC}
A template-based tracker simply uses the target feature (i.e., the feature of the target) as its target model. Intuitively, it means that the confidence of the test patch being the target is high when it is similar to the target feature and low otherwise. Due to the simplicity of the target model and the learning algorithm (i.e., an identity function), the performance of template-based tracker depends heavily on its features. 

SiamFC\cite{bertinetto2016fully} is a template-based tracker which achieves good performance with an offline learned feature extractor. The feature extractor is learned under a two branch structure with millions of image pairs sampled from videos. Each image pair contains a target patch and a test patch. The feature of the target patch is computed as the linear target model. The learning objective is that the confidence of a test patch containing the target is high while the confidence of background being low.

Interestingly, SiamFC does not have a model update module. In other words, given the target feature in the first frame $\bar{x}_1$ (we will use $\bar{x}$ for target feature and $x$ for the feature of any image patch), the online training set contains only the target feature $T_t = \{\bar{x}_1\}$ and the learning algorithm simply takes the target feature as the target model $\theta_{t+1} = u(T_t) = \bar{x}_1$. 

The detection process of SiamFC is equivalent to computing the similarity by sliding the target model $\theta$ over the search feature $z$ and then outputting the position with the largest response. To facilitate fast detection, the feature extractor is designed to be fully convolutional and the similarity metric is simply inner product. Therefore, the detection process can be readily implemented via cross-correlation $\star$ (or convolution in the neural network literature):
\begin{equation} \label{eq:template-detect}
	\begin{aligned}
		p = \arg\max_{p} \theta \star z \enspace .\\
	\end{aligned}
\end{equation}  

\subsection{Correlation Filter-Based Tracker: CFNet}
Correlation filters are one representative example of linear target model that have shown superior performance and gained a lot of popularity. The key factor behind the success of correlation filters is an efficient learning algorithm that is able to handle tens of thousands of circulant training samples. Correlation filter-based trackers have been improved in various aspects since the seminal work of \cite{bolme2010visual}. For simplicity, we use the basic formulation and follow the setup in CFNet\cite{valmadre2017end}.

In correlation filter-based tracker, given one base sample of the target feature $\bar{x}$, various virtual samples are obtained by cyclic shifts. The label of the base sample is $1$ while the labels of the virtual samples follow a Gaussian function depending on the shifted distance. Given the samples and the corresponding labels, a ridge regression problem is solved efficiently in the Fourier domain making use of the property of circulant matrix\cite{gray2006toeplitz}. In this work, we will simply use $\mathrm{CF}(\cdot)$ to denote the algorithm of learning the correlation filter from \emph{one base sample with multiple feature channels}. Please refer to \cite{bolme2010visual, henriques2015high} for more details about $\mathrm{CF}(\cdot)$.

The detection of correlation filter-based tracker is typically conducted in the Fourier domain. However, we convert the learned filter back to the spatial domain following CFNet. In this way, the detection process is the same as the SiamFC tracker. Moreover, it frees the learned updater from dealing with complex values.

\subsection{Two Baseline Model Update Methods}
For linear target model based trackers, there are two commonly used model update methods. We briefly introduce these two baseline methods in this subsection.  

\textbf{EMA-based model update.} Let $g: \mathcal{X} \to \Theta$ be the algorithm of learning a linear target model using a \emph{single} target feature. For template-based tracker, it is the identity function. For correlation filter-based tracker, it is $\mathrm{CF}(\cdot)$. Typically, the online training set contains the target feature at each frame, i.e., $T_t = \{\bar{x}_1, \bar{x}_2, ..., \bar{x}_t\}$. It is clear that $T_t = T_{t-1} \cup \{\bar{x}_t\}$. Given the online training set $T_t$, the updated target model via EMA is 
\begin{subequations}\label{eq:exponential-moving-average}
	\begin{align}
    \theta_{t+1} &= u(T_t)  \\
                 &= (1 - \alpha) u(T_{t-1}) + \alpha g(\bar{x}_t) \\
                 &= (1 - \alpha)\theta_t + \alpha \theta'_t
	\end{align}
\end{subequations}
where $\alpha$ is the learning rate that controls the rate of adaptation, $\theta'_t = g(\bar{x}_t)$ is the candidate target model at time $t$. Note that this update method only requires the last target model $\theta_t$ and the current target feature $\bar{x}_t$. Hence, there is no need to explicitly collect a large online training set.

EMA-based update is widely adopted in correlation filter-based trackers because it is easy to implement and efficient in terms of both memory consumption and computational complexity. Moreover, EMA-based update is the update rules of correlation filters derived for the multiple base samples with \emph{single channel} case \cite{bolme2010visual} and can be quite effective even for samples with multiple channels. However, in general, EMA-based update is ad hoc and only marginally improves the template-based tracker as shown in our experiments (Section~\ref{sec:experiments}).

\textbf{SGD-based model update.} Besides EMA, another update method is to collect the online training set using samples from the search space of detection, i.e., $T_t = \{(z_1, y_1), (z_2, y_2), ..., (z_t, y_t)\}$ where $z_t$ is the feature of the search image at frame $t$ and $y_t$ is the corresponding label map. The update process based on SGD is
\begin{equation}
  \theta_{t+1} = \theta_{t} - \alpha \frac{\partial l_{T_t^b}(\theta)}{\partial \theta}
\end{equation}
where $l_{T_t^b}(\theta)$ is a differentiable objective function with variables $\theta$ over a batch of samples $T_t^b$ from the online training set, and $\alpha$ is the learning rate. 

Typically, a fixed size of online training set is maintained by dropping the earliest samples when the capacity is exceeded. However, it still contains thousands of samples. Moreover, we show only one iteration of SGD update for brevity while it generally requires dozens of iterations to take effects. These computational burdens of the SGD-based update method hinder its practical usage.

\begin{figure}
	\centering
	\includegraphics[width=1\columnwidth]{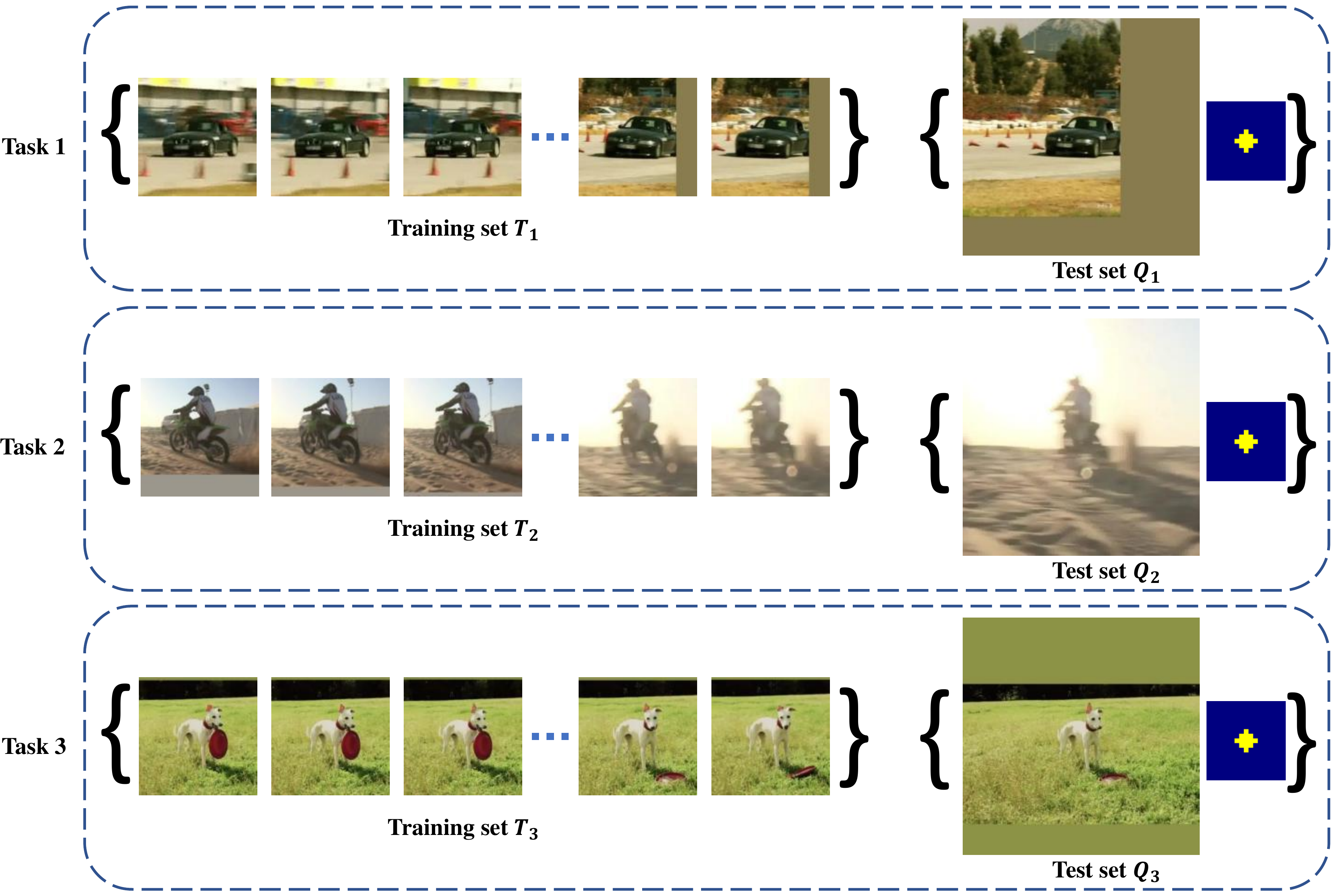}
	\caption{Tasks for learning the model updater during offline training. Each task consists of a training set and a test set. The model updater should learn from the training set and generate a target model which is tested on the test set.}
	\label{fig:datasets}
\end{figure}

\begin{figure*}
	\centering
	\includegraphics[width=1.7\columnwidth]{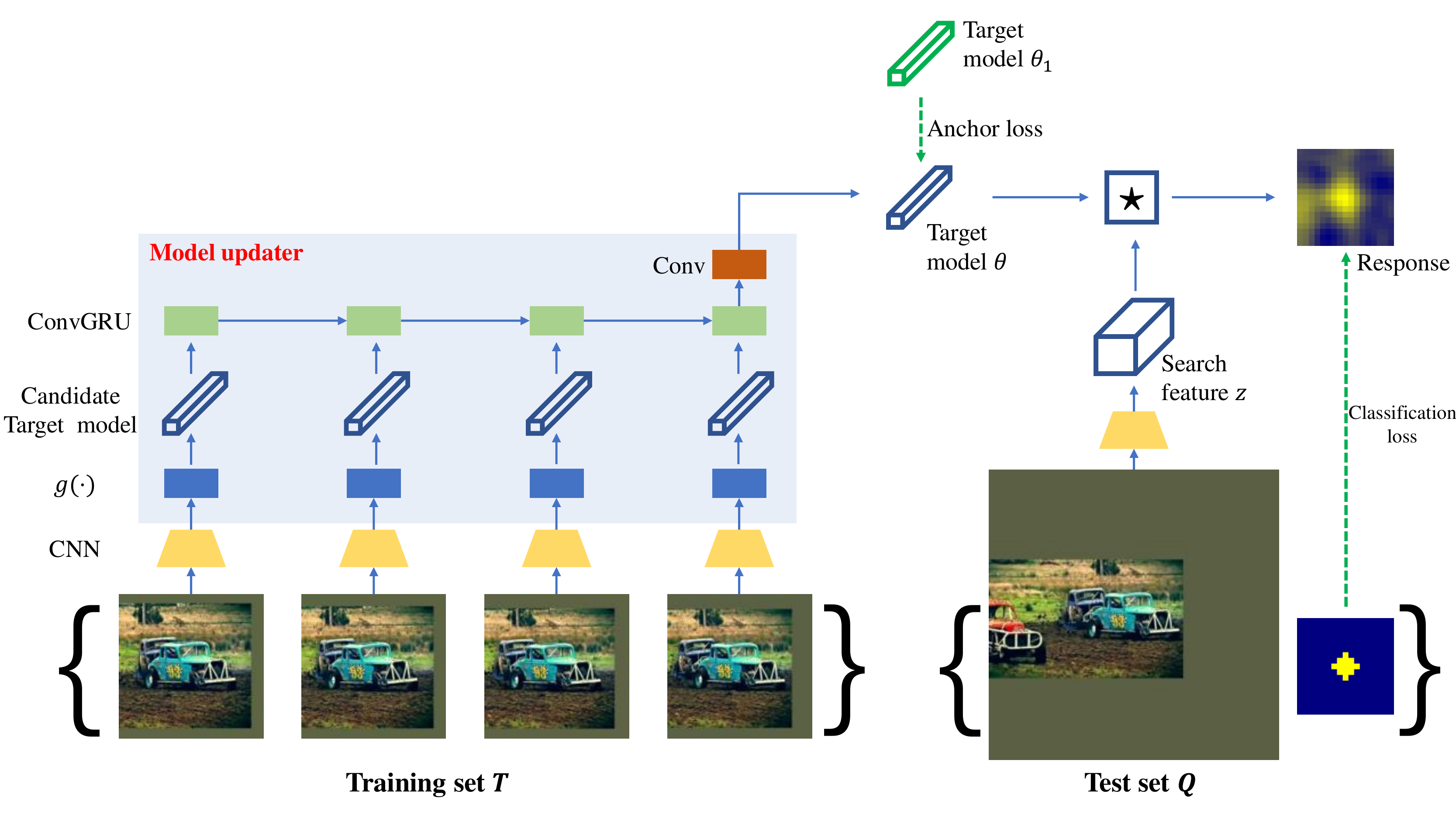}
	\caption{The framework of learning to update during offline training. Given the training set $T$ with image patches of a car, the target model $\theta$ is updated by the recurrent model updater. The target model is tested on the test set $Q$ to obtain the classification loss. An anchor loss is also added to improve generalization. The model updater is learned by optimizing the anchor loss and the classification loss. During tracking, the model updater is fixed. An online training set is gathered along with tracking as $T$ and the target model is updated by the model updater and applied to subsequent frames.}
	\label{fig:model-updater-framework}
\end{figure*}

\section{Learning to Update}\label{sec:learning-to-update}
In this section, we first formulate the model update problem in the meta-learning framework and then introduce the proposed model updater.

\subsection{Model Update as Meta-Learning}
\begin{table}
	\centering
	\caption{Correspondence between the meta-learning and learning to update. Best viewed by zooming in the electronic version.\label{tab:correspondence}}
	\resizebox{1\linewidth}{!}{%
	\begin{tabular}{l|l|l}
		\hline
		\textbf{meta-learning} & \textbf{learn to update} & Explanation\\
		\hline
		meta-training & offline training & The process of learning a meta-learner (model updater)\\
		\hline
		meta-test & online training & The process of applying the 
		meta-learner (model updater)\\
		\hline
		meta-learner & model updater & The model that takes as input an training set and \\ 
		 & & outputs a learner (target model)\\ 
		\hline
		a task & a task & An example used for meta-training (offline training) \\
		& & the meta-learner (model updater),\\ 
		& & consisting of a training set and a test set \\
		\hline
		training set & training set & The dataset for learning a learner (target model)\\
		\hline
		test set & test set & The dataset for testing the learner (target model) \\
		\hline
		learner & target model & The model learned from the training set \\
		\hline
	\end{tabular}
	}
\end{table}

Meta-learning is about learning from large numbers of tasks during meta-training to quickly learn a \emph{new} task at meta-test time.
Each task consists of a training set and a test set. The meta-learner (i.e. the model to be obtained via meta-training) takes as input the training set of a task and outputs a learner which should perform well on the corresponding test set. The aim of meta-learning is to learn a meta-learner during meta-training such that the meta-learner can quickly learn a good learner at meta-test time, so meta-learning is also known as learning to learn.

The uniqueness of meta-learning is that, at meta-test time, the meta-learner should \emph{quickly learn a new concept} using examples from the training set of the concept. An example is to learn a classifier to differentiate ``Apple" and ``Pear" based on examples of each category where these two categories never appear during meta-training. This resembles model update for object tracking since the tracker should quickly update the target model to accommodate an object that does not appear during offline training\footnote{In fact, during offline training, the ``human'' category never appears whereas humans appear often in the tracking benchmarks.}.

In the context of meta-learning for target model update, we aim to learn a meta-learner (model updater) from large numbers of offline videos during meta-training (offline training). Each task is to learn the learner (target model) from the training set to locate the target at test set. The correspondence between learning to update and meta-learning is summarized at Table~\ref{tab:correspondence}.

During offline training, given the training set $T$ and the test set $Q$ of a task constructed from the offline videos $(T, Q) \in \mathcal{V}$, the model updater computes target model $\theta = u_\phi(T)$. Let $l(\theta, Q)$ be the loss of the target model on the test set $Q$ of a task. The model updater $\phi$ is learned by minimizing the following loss: 
\begin{equation}
	\begin{aligned} \label{eq:meta-loss}
		\mathcal{L}(\phi) &= \sum_{(T, Q) \in \mathcal{V}} l(u_{\phi}(T), Q) \enspace .
	\end{aligned}
\end{equation}

\subsection{The Training Set and the Test Set}
We now describe how to construct the training set and the test set of a task for learning to update.

Given a video from the offline videos and the corresponding target positions (width, height, center position) at each frame, we first normalize the scale variations of targets by scaling the image with factor $s$ such that $s(w + 2p) \times s(h + 2p) = A$ where $w, h$ are the width and height of the target in the image, $p = \frac{w + h}{4}$ is the context margin and $A = 127 \times 127$ is the desired target size after scaling.

A subset of $N$ image frames are sampled from the scaled images while keeping the temporal order. The first $N - 1$ frames are cropped at the center of the target, with size $127\times 127$, and used as the training set of the target. The last frame is also cropped at the center of the target, with size $255\times 255$, and used as the test set. Note that we use a larger image patch in the test set since both of our base trackers are translation equivalent and the $255\times 255$ image can be seen as a set of $127\times 127$ images. Please refer to Fig.~\ref{fig:datasets} for several examples of training set and test set.

Cropped images are then embedded by the feature extractor. Finally, we have the training set $T = \{\bar{x}_1, \bar{x}_2, ..., \bar{x}_{N-1}\}$ where $\bar{x} \in \mathbb{R}^{m\times m \times d}$ is the feature of the target. The test set $Q = \{(z, y)\}$ where $z \in \mathbb{R}^{n\times n \times d}$ is the feature of the search image and $y\in \{-1, +1\}^{(n - m + 1)\times (n - m + 1)}$ is the corresponding label map.
Features have spatial size $m$ or $n$ and channel size $d$.

\subsection{Instantiation of The Learned Updater}
A model updater takes as input a training set $T = \{\bar{x}_1, \bar{x}_2, ..., \bar{x}_{N-1}\}$ and outputs the updated target model $\theta$, i.e., $\theta = u(T)$. The design of the updater includes several preferable properties: 1) supporting training set with variable size; 2) incremental update, i.e., during tracking, the target model is updated based on existing values instead of learning from scratch; 3) memory and computational efficiency.

In this work, we propose a RNN-based updater that satisfies all these properties. Concretely, our updater follows a three-step procedure.

\textbf{Step 1: Project from feature space to model space.} We first project each target feature in the training set into the model space by $\theta' = g(\bar{x})$, 
where $\theta' \in \mathbb{R}^{m\times m \times d}$ is the candidate target model and $g(\cdot)$ is the algorithm of learning a linear target model using a single target feature. In particular, $g(\cdot)$ is the identity function for SiamFC and is the $\mathrm{CF}(\cdot)$ function followed by a center cropping function for CFNet.

\textbf{Step 2: Aggregate target information.} We use RNN to summarize the training set into a single tensor. For simplicity, gated recurrent unit (GRU) is adopted \cite{cho2014learning}. We find GRU achieves better performance than the Long-short term memory (LSTM\cite{hochreiter1997long}) in the ablation study. To preserve the spatial dimension, we extend the original GRU formulation to Convolutional GRU (ConvGRU) by replacing all matrix multiplications with convolutions.

\textbf{Step 3: Generate target model.} Given the last hidden state of the ConvGRU, one convolutional layer is used to generate the target model.

By adopting RNN, our updater is able to handle training set with variable size. Moreover, the target model is updated in an incremental manner. To make things clear, during tracking, denote $T_{t-1} = \{\bar{x}_1, \bar{x}_2, ..., \bar{x}_{t-1}\}$ as the online training set at time $t-1$. After model update, we obtain the hidden state $h_{t-1}$. At time $t$, the online training set is updated with a new example $\bar{x}_t$, i.e., $T_t = T_{t-1} \cup \{\bar{x}_t\}$. Since the first $t-1$ examples are unchanged, we can simply reuse $h_{t-1}$ and get $h_t = \text{ConvGRU}(h_{t-1}, \bar{x}_t)$. Moreover, with incremental update, we can avoid explicitly storing and manipulating a large online training set during tracking since our updater only needs $h_{t-1}$ to generate $h_t$ which saves a large amount of memory.

Until now, the updater is restricted to use $N-1$ target features as the training set and one search image feature as the test set. Note that given a sequence of $N - 1$ target features, our updater can readily compute $N - 1$ target models at each time step. Therefore, given a video with $N$ images, we can construct various training sets with length $1, 2, ..., N - 1$ and the computation for model update can be shared.

\subsection{The Learning Objective}
Given the updated target model $\theta$, we need a ``goodness'' measurement of the target model which in turn indicates how good the updater is and thus enables optimization. 

\textbf{Classification Loss}. Following the meta-learning framework, the updated model is evaluated on the test set $Q = \{(z, y)\}$. Using the normalized logistic loss for classification, we have
\begin{equation}
  l_c(\theta; Q) =  \frac{y(-\ln\sigma(\theta \star z)) + (1 - y)(-\ln(1 - \sigma(\theta \star z)))}{(n - m + 1)\cdot (n - m + 1)}. 
\end{equation}

\textbf{Anchor Loss}. At a first glance, it would seem that classification loss is all we need to train the updater. However, model update faces the intrinsic problem: \emph{the stability-plasticity dilemma}, i.e., model update should be stable with respect to noise and flexible to assimilate new information. With only classification loss, since $z$ is close to $\bar{x}_{N-1}$, the updater will adopt an aggressive update strategy and store new information brought by $\bar{x}_{N-1}$ as much as possible. The problem is that $\bar{x}_{N-1}$ is not always reliable during tracking and thus the updater trained with only classification loss is prone to small errors. 

One effective method that is validated by the literature is to use the target model at the first frame as an anchor point \cite{matthews2004template, guo2017learning}. Given inadequate training data, such an anchor point is hard to learn without regularization. Therefore, we design an anchor loss, which penalizes the updater when the updated target model drifts away from the initial target model:
\begin{equation}
	\begin{aligned}
		l_a(\theta; \theta_1) &= \frac{1}{m\cdot m \cdot d}||\theta - \theta_1||_2^2 \enspace .
	\end{aligned}
\end{equation}
where the loss is normalized by the number of target model parameters $m\cdot m \cdot d$.

\textbf{Total Loss}. Classification loss and anchor loss are linearly combined for measuring a target model: 
\begin{equation}
	\begin{aligned} \label{eq:target-loss}
		l(\theta; Q) &= (1 - \lambda) l_c(\theta; Q) + \lambda l_a(\theta; \theta_1) \enspace ,
	\end{aligned}
\end{equation}
where $\lambda$ is the combination factor. Successful learning of the updater should maintain a good balance between the classification loss and the anchor loss.

The total loss of the updater $u_{\phi}(\cdot)$ with learnable parameters $\phi$ is then the loss of the target model in the offline training set by inserting Eq.~\ref{eq:target-loss} into Eq.~\ref{eq:meta-loss}.

\subsection{Practical Techniques for Effective Learning}\label{sec:effective-techniques}
Our learned updater collects target information based on RNNs, which are well known for modeling sequential/temporal variations. However, the problem of limited offline training videos has to be addressed before it unleashes the power. We describe several techniques based on the nature of model update that turn out to be effective.

\textbf{Modeling long-term variation by truncated backpropagation}. Typically, a subset of $N$ frames are sampled from the original video for RNN training. For convenience, we define \emph{maximum modeling length} of an algorithm to be the largest length of all sampled sequences during training, where the length of a sampled sequence stands for the distance counted by \emph{\#(frames) in the original video} between the first and the last sampled frame. Since videos have hundreds of frames to track, it is desirable for RNN-based models to learn long-term dependency with a large maximum modeling length. However, training with long sequences is computationally demanding and may incur the vanishing gradient problem. To avoid such a problem, existing RNN-based trackers sample relatively few frames sparsely from the original video. For example, \cite{yang2017recurrent} samples training sequences with 10 frames and large frame interval (30 frames on average). Such a sparse sampling strategy, however, enlarges the target variations between sampled frames, which is more difficult to learn. We conjecture that these trackers are disadvantaged by such limitations. 

Contrarily, to train our learned updater, we sample training sequences with as many as 150 frames and small frame interval ($\leq 2$ frames)\footnote{In this sense, the maximum modeling length of our method here is 300 frames.}. To handle the aforementioned problem of training with long sequences, instead of backpropagating all the way to the first frame, we adopt truncated backpropagation. This method processes training frames one timestep at a time, and every $H$ timesteps, it runs backpropagation through time (BPTT) for $H$ timesteps. $H$ is called the \emph{unroll length}.

\textbf{Matching training and testing behavior by estimated target position}. During offline training of the learned updater, training set $T$ of the training video needs to be generated. In our case, the online training set contains the target features at each frame $T_t = \{\bar{x}_1, \bar{x}_2, ..., \bar{x}_t\}$. The extraction of the target features depends on the target position which can only be estimated during tracking. RNNs often take as input the groundtruth during training, which is known as teacher forcing. However, as noted in \cite{bengio2015scheduled}, this causes discrepancy between training and testing and hampers the performance of RNNs. In this work, we always use the target position during training that is inferred by the target model to keep in line with testing.

\textbf{Reducing overfitting by interval update}. As noted in \cite{Danelljan2016ECOEC}, instead of updating the target model in every frame, it is beneficial to apply a sparser update scheme. We adopt this simple strategy by updating the target model every $M$ frames. Note that \emph{the hidden state of RNN is updated in every frame} though. The reason for the improved performance is that by updating after a certain timesteps, the learned updater can make more informed update decision, and therefore reduces the risk of overfitting to current training samples.

\section{Experiments}\label{sec:experiments}
\subsection{Implementation Details}
\textbf{Training data}. The feature extractor and learned updater are trained offline on the ILSVRC 2015 Object Detection from Video dataset (Imagenet VID) \cite{russakovsky2015imagenet}. Imagenet VID contains 4417 videos and each video has about 2 object tracks on average, adding up to 9220 tracks. Tracks are annotated with bounding boxes in each frame and contain about 230 frames on average. This differs from another large-scale video dataset, namely Youtube-BB \cite{real2017youtube} where objects are annotated every 30 frames and each track has about 15 annotated frames. We find that small frame interval is important for training the learned updater, and thus, we use Imagenet VID instead of the much larger Youtube-BB. However, it is worth investigating effective ways to make use of Youtube-BB for object tracking. 

Image sequences are sampled from tracks as training samples for the learned updater. We use bucketing \cite{khomenko2016accelerating} (i.e., an RNN training technique which batch together sequences of similar lengths for efficiency) to handle sequences of different length. Multiples of the RNN unroll length are used as the bucket sizes. For example, bucket sizes of 25, 50, ..., 125, 150 are used for fast experimentation where 25 is the RNN unroll length. For tracks that are longer than the largest bucket size (e.g., 150), we sample a portion of the tracks with small frame interval (e.g., 1 or 2). For short tracks, we lengthen these tracks by duplicating frames or simply drop these tracks according to probabilities that are proportional to the track length. Images are preprocessed according to its base tracker SiamFC \cite{bertinetto2016fully} and CFNet \cite{valmadre2017end}. Particularly, these two base trackers use the same preprocessing procedures to crop and resize images such that targets are at the image center and take up 127 x 127 pixels together with context.

\textbf{Architecture}. For template-based tracker, we use the same modified Alexnet architecture in \cite{bertinetto2016fully} for the feature extractor. 
 For correlation filter-based tracker, we use the 3 layer CNN feature extractor in \cite{valmadre2017end}, which is trained following the procedures in \cite{bertinetto2016fully}. All of our experiments stack two convolutional GRU layers, where convolution operations have kernel size 3 with zero padding to preserve spatial dimension. For template-based tracker, each convolutional GRU layer has 192 units while for correlation filter-based tracker, each has 64 units\footnote{The number of hidden units are set by searching from 32 to 384 with step size 32. We empirically find that setting the number of hidden units close to the channel size of the target model performs well.}. One convolutional layer with kernel size 3 is used to generate the updated target model based on the hidden states, which takes as input the concatenated states of the two convolutional GRU layers and outputs corresponding target models. Dropout \cite{zaremba2014recurrent} and layer normalization \cite{ba2016layer} are added in each convolutional GRU layer to avoid overfitting.

\textbf{Optimization}. Learned updaters are trained over 60 epochs, each epoch consists of 8309 image sequences. Gradients are computed using mini-batches of size 8, which are used by the Adam optimizer \cite{kingma2014adam}. Learning rate is fixed to be 1e-4. Weight decay is 5e-4. 

\textbf{Hardware and software specifications.} The speed measurements of our trackers are performed on a computer with an Intel Core i7-5930K Haswell-E 6-Core 3.5GHz CPU and a GeForce GTX 1080 GPU. Our trackers are implemented in TensorFlow \cite{tensorflow2015-whitepaper}, which is compiled with CUDA 8.0 and cuDNN 6.0. 

\textbf{Postprocessing.} We adopt the same strategy as our base trackers for penalizing large displacement and handling scale variations. Specifically, a cosine window is added to the response map to penalize the large displacement. For scale estimation, three search patches with different scales are extracted and the current scale is calculated by interpolating the newly predicted scale with a damping factor.

\subsection{Benchmarks and Evaluation Protocols}
\textbf{OTB}. The OTB benchmark contains three subdatasets: OTB-2013, OTB-50 and OTB-100, each of which consists of 51, 50 and 100 natural image sequences, respectively. The standard evaluation metric on OTB is the area under curve (AUC) of the threshold-success rate curve which represents the success rates at different thresholds. For each frame, the overlap (intersection over union) between the predicted target bounding box and groundtruth is computed. The success rate at a given threshold corresponds to the fraction of frames that has overlap no less than the given threshold. 

\textbf{VOT}. The VOT benchmarks are a collection of tracking challenges held on a yearly basis starting from 2013. We use three recent benchmarks: VOT-2015, VOT-2016 and VOT-2017. Unlike OTB, which lets the tracker run until the end of the image sequence, VOT focuses on short-term tracking (no redetection is required) and resets the tracker once it drifts away from the target. The primary measure is the expected average overlap (EAO), which reflects the similar property as AUC. VOT-2017 also introduces a new ``realtime challenge", where a tracker is constantly receiving images in realtime speed and if the tracker does not respond after a new frame becomes available, the last bounding box predicted by the tracker is reported for the current frame. 

\subsection{Ablation Study}

\begin{table*}[]
	\caption{\textbf{Ablations} on the OTB-2013 dataset using template-based tracker. Only color images are adopted during offline training. RNNs are unrolled 25 steps by default. \label{tab:ablations}}

	\subfloat[\textbf{Effective techniques:} Cross mark means the technique is removed. RNNs are unrolled 25 steps by default unless large unroll length (50) is adopted. Image sequence are sampled by default with small frame interval ($\leq 2$). By removing the technique, we use large frame interval ($\geq 10$). \label{tab:technique-ablations}]{
		\begin{tabular}{r|ccccc}
			\hline
			interval update?        		& \cmark 		& \cmark 		& \cmark 		& \cmark 		& \xmark    \\
			estimated target position?      & \cmark 		& \cmark 		& \cmark 		& \xmark 		& \cmark    \\
			small frame interval?   		& \cmark 		& \cmark 		& \xmark 		& \cmark 		& \cmark    \\
			large unroll length? 	  		& \cmark 		& \xmark 		& \xmark 		& \xmark 		& \xmark    \\
			\hline
			AUC (\%)                     	& 64.4  		& 63.6  		& 63.0 	 		& 62.6 			& 61.7	    \\
			\hline
		\end{tabular}}\hspace{3mm}
	\subfloat[\textbf{RNN configurations}: For example, ConvGRU-192x2 denotes stacking 2 layers of ConvGRU, each with hidden size 192.\label{tab:rnn-ablations}]{
		\begin{tabular}{l|c}
			\hline
			configuration & AUC (\%) \\
			\hline
			ConvGRU-192x1 & 62.1 \\
			ConvGRU-192x2 & \textbf{63.6} \\
			ConvGRU-192x3 & 63.3 \\
			\hline
			ConvGRU-128x2 & 62.8 \\
			ConvGRU-192x2 & \textbf{63.6} \\
			ConvGRU-256x2 & 63.4 \\
			\hline
			ConvLSTM-128x2 & 62.2 \\
			ConvLSTM-192x2 & 62.8 \\
			ConvLSTM-256x2 & 62.6 \\
			\hline
		\end{tabular}}\hspace{3mm}
		\subfloat[\textbf{Joint training:} F: feature extractor, M: model updater, F + M: joint training of feature extractor and model updater. \label{tab:joint-training}]{
			\begin{tabular}{l|l|c}
			\hline
			& Description & AUC (\%) \\
			\hline
			scheme 1 & stage 1: F + M & 51.5 \\
			\hline
			\multirow{2}{*}{scheme 2} & stage 1: F & \multirow{2}{*}{56.2}  \\
									   & stage 2: F + M & \\
			\hline
			\multirow{3}{*}{scheme 3} & stage 1: F & \multirow{3}{*}{63.1} \\
					 & stage 2: M & \\
					 & stage 3: F + M & \\
			\hline
			\multirow{2}{*}{Ours} & stage 1: F & \multirow{2}{*}{\textbf{63.6}} \\
				 & stage 2: M & \\
			\hline
			\end{tabular}
		}

		\subfloat[\textbf{Anchor loss:} Tracker accuracy as the combination factor $\lambda$ varies during updater training. ``lu" is our learned updater. \label{fig:anchor_loss_weight}]{
			\includegraphics[width=0.63\columnwidth]{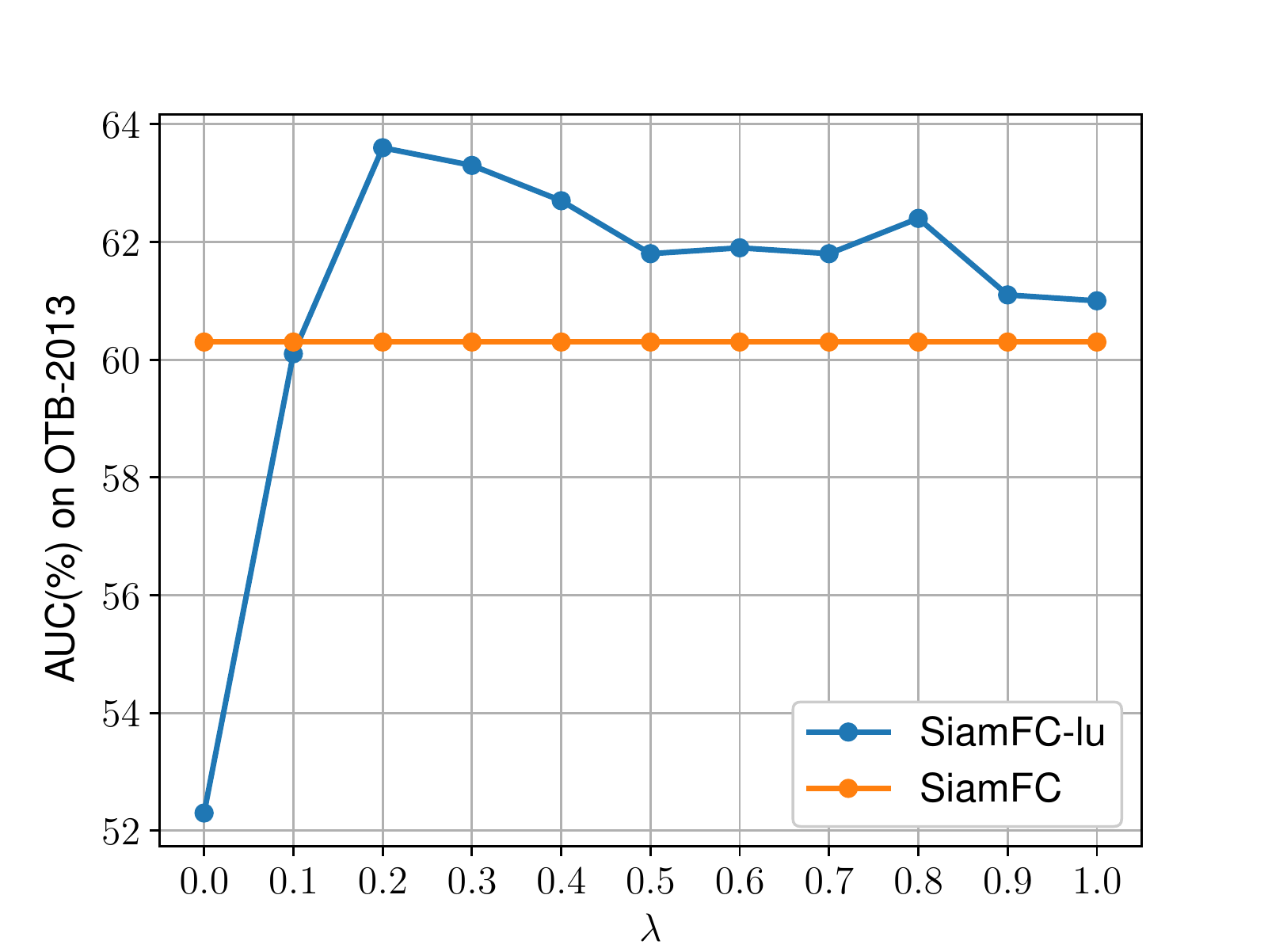}
		}\hspace{3mm}
		\subfloat[\textbf{Maximum modeling length of feature extractor:} Tracker accuracy as the maximum modeling length of the feature extractor varies during offline training, with and without the learned updater. \label{fig:SFC_training_sequence_distance}]{
			\includegraphics[width=0.63\columnwidth]{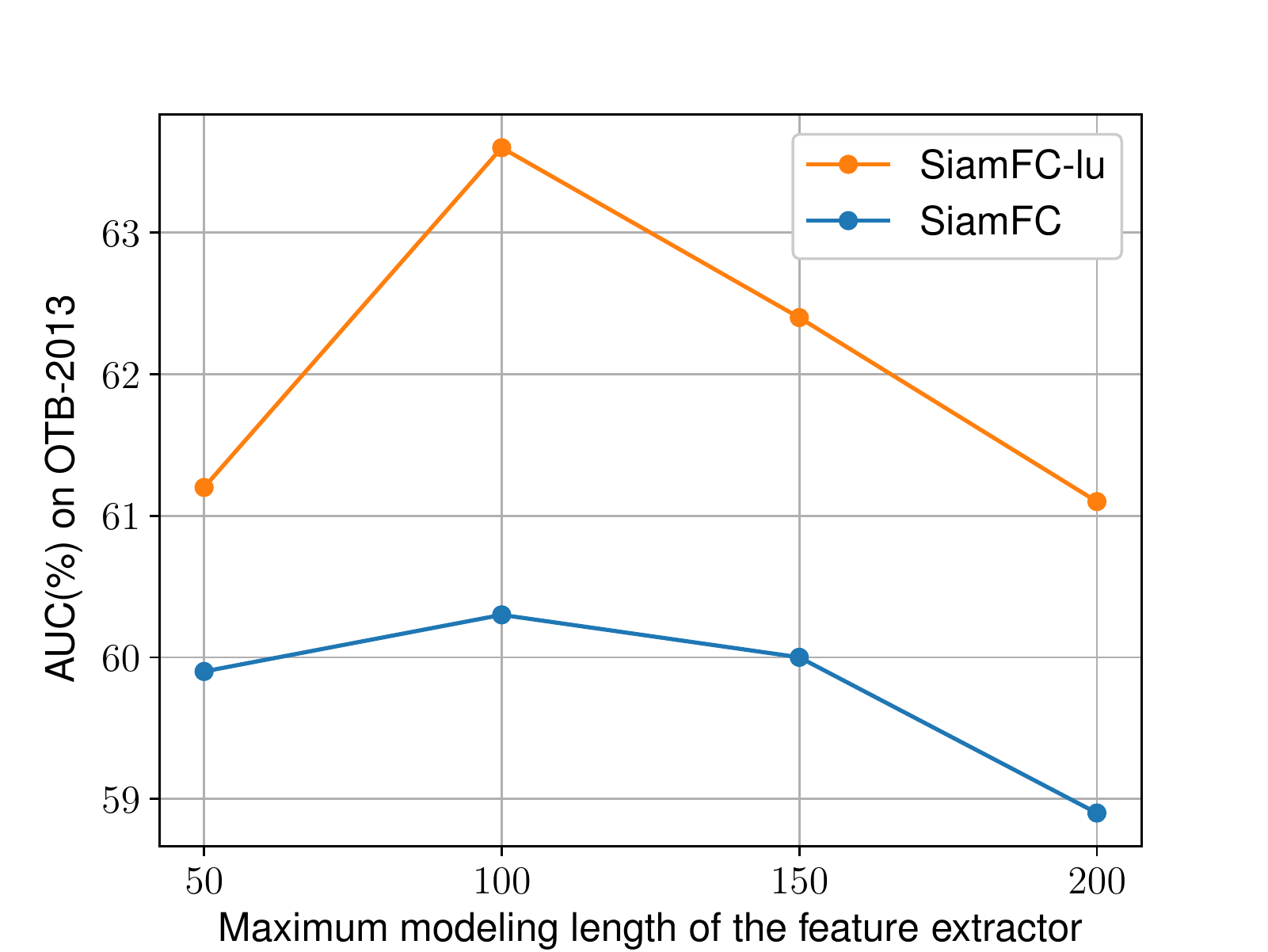}
		}\hspace{3mm}
		\subfloat[\textbf{Maximum modeling length of model updater:} Tracker accuracy with varied maximum modeling length and unroll length of the model updater during offline training. \label{fig:training_sequence_length}]{
			\includegraphics[width=0.63\columnwidth]{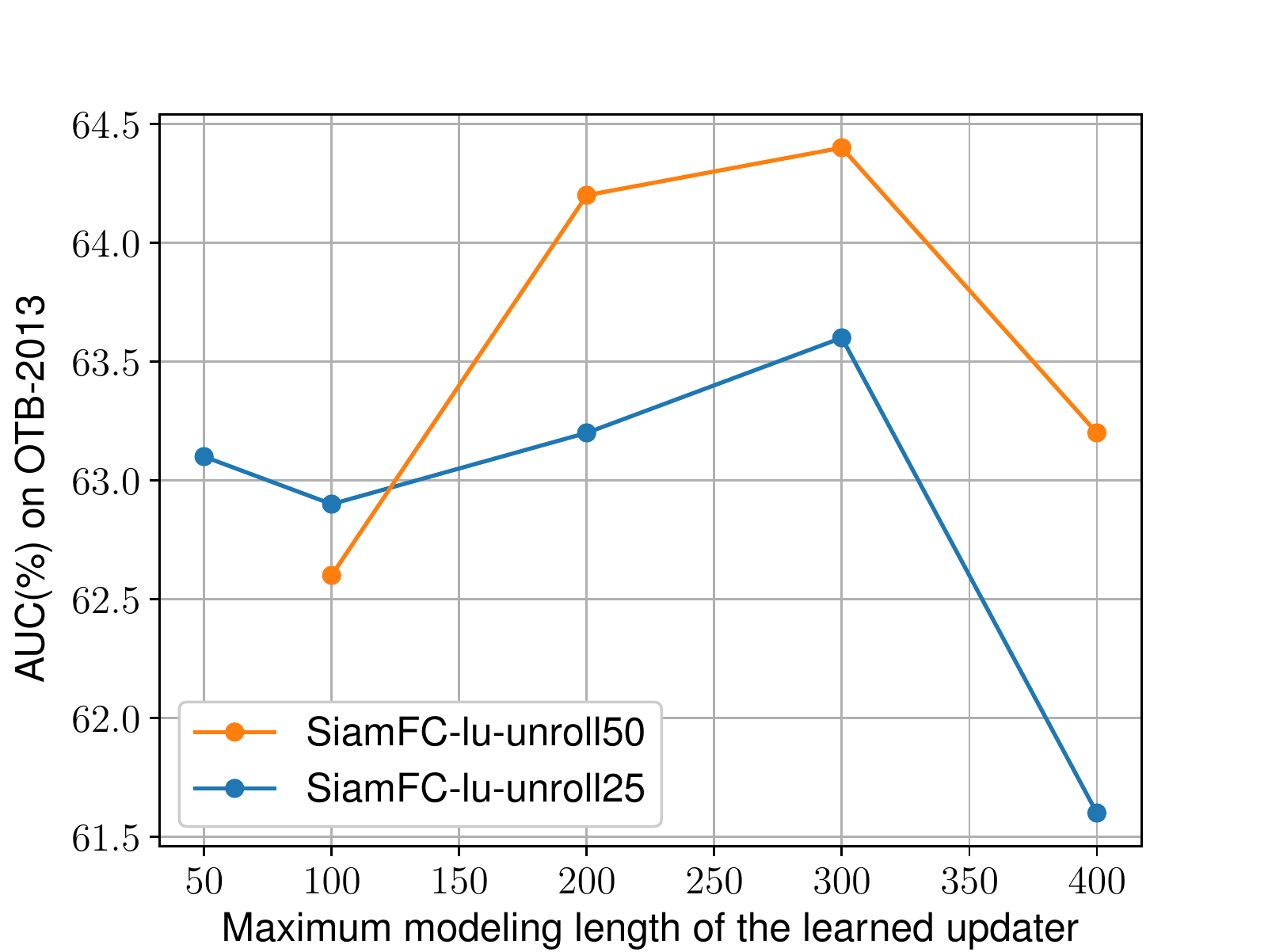}
		}\hspace{3mm}
\end{table*}

We validate the effectiveness of various designs of our learned updater based on the template-based tracker. OTB-2013 is used for ablation study. All experiments use the same configurations except the components that are examined. Although larger unroll length typically gets better results, in the ablation study section, we use unroll length 25 for fast experimentation which is also the default configuration unless larger unroll length is needed. Moreover, only color images are adopted during offline training to prevent benchmark-specific choices\footnote{OTB-2013 contains both color and grayscale videos. Therefore, models trained with both color and grayscale images typically perform better in this benchmark.}. All models are trained using our implementation including SiamFC. Results are summarized in Table~\ref{tab:ablations}.

\textbf{Anchor loss is crucial for successful learning.} To overcome the stability-plasticity dilemma, we propose the anchor loss which penalizes large variations of the updated target model. To minimize the anchor loss, one straightforward strategy would be \emph{no update at all}. However, besides the anchor loss, the learned updater is also constrained by the classification loss which encourages the update of the target model to keep up with target variations. It is of interest to investigate how the interplay between these two losses affects the performance of the learned updater.

Our learned updaters trained with different combination factors $\lambda$ are shown in Table~\subref{fig:anchor_loss_weight}. We also include the results of the no-update baseline (i.e., the setup in SiamFC\cite{bertinetto2016fully}) for reference. When $\lambda = 0$ (i.e., no anchor loss), our learned updater is not constrained to generate target models that are consistent with the initial target model and can quickly drift away. The performance is even worse than not updating at all. As $\lambda$ increases, the learned updater gets better and reaches the peak at 0.2. Further increasing the anchor loss weight diminishes the possible target model choices of the learned updater and the performance drops. Note that even without the classification loss (i.e., $\lambda$ = 1), our learned updater outperforms the baseline with no update. The reason is that the learned updater is given the initial target model only once. After that, it constantly receives new target model information and the hidden states of RNN inevitably stores new information due to the soft store operations. It is this new information that helps tracking.

\textbf{Learned updater is orthogonal to feature extractor}. Our template-based tracker is based on \cite{bertinetto2016fully}, which aims to learn an invariant and discriminative feature extractor such that model update is not necessary. Two interesting questions are: 1) how much variation the Siamese network is able to learn, and 2) how the learned feature extractor affects our learned updater. We answer these questions from the perspective of training samples of feature extractor. Every training sample of the Siamese network contains two image patches from two frames of a video. These two patches are both centered on the target and at most $K$ frames apart, and therefore the maximum modeling length is $K$ according to the definition in Section~\ref{sec:effective-techniques}. We investigate the effects of different $K$, and train an updater based on each feature extractor. Results are shown in Table~\subref{fig:SFC_training_sequence_distance}.     

As can be seen, 1) Siamese network is able to capture the variations of target within 100 frames, but has difficulties learning beyond this limit. 2) Although the feature extractor is trained with the objective of invariance and does not require model update, our learned updater consistently improves the base trackers. Moreover, the improvement is \emph{positively correlated} with the performance of the feature extractor.

\textbf{Train longer, generalize better.} Model update is a process that typically spans hundreds of frames. We investigate the effects of training with long image sequences (large maximum modeling length). Table~\subref{fig:training_sequence_length} shows the tracking results of the learned updater trained with different maximum modeling length and unroll lengths. By increasing the maximum modeling length from 100 to 300, the learned updater monotonically gets better results. Since modeling long term dependencies is still a challenge for RNN, the performance degenerates when it reaches 400. In conclusion, 1) large maximum modeling length helps the learned updater to generalize better if it is within the modeling capabilities of RNN; 2) large unroll length is still helpful under truncated backpropagation to model long term dependencies.

Moreover, it is worth mentioning that the performance of the Siamese network decays as the maximum modeling length is over 100 (as shown in Table~\subref{fig:SFC_training_sequence_distance}). We have tried to increase the number of neurons in Siamese network, but it does not help. Contrarily, our learned updater can handle longer sequences (e.g., 300 frames). It can be inferred that it is difficult to extract invariant features to handle long-term target variation, whereas learning an updater to adapt the target model gradually is relatively easier.

\textbf{Practical techniques are helpful}. We demonstrate the effectiveness of the various practical techniques introduced in Section~\ref{sec:effective-techniques} by removing one component at a time. Note that small unroll length is used by default instead of large unroll length for fast experimentation. The results are summarized in Table~\subref{tab:technique-ablations}. As shown in the table, every component has contributed to the final performance. Among these techniques, interval update plays such an important role that the performance drops about 2\% once removed. This is in accordance with the findings in \cite{Danelljan2016ECOEC}.

\textbf{Ablation on the RNN.} Table~\subref{tab:rnn-ablations} shows the results of RNNs with different configurations of cell unit (ConvGRU or ConvLSTM), number of stacked RNN layers and the hidden state size. As can be seen, ConvGRU consistently outperforms ConvLSTM which may be related to the observation that the best performing cell unit is task-dependent\cite{chung2014empirical}. We also observe that the AUC largely increases from 62.1\% to 63.6\% by stacking two layers of ConvGRU. The performance is not sensitive to the size of the hidden state as long as it is in a reasonable range (192\textasciitilde256).

\textbf{Joint training is difficult.} The whole system (including feature extractor and model updater) can be trained jointly. We have tried several schemes for joint training, of which the results are summarized in Table~\subref{tab:joint-training}. Interestingly, it is best to train feature extractor and model updater separately - first train the feature extractor without model updater, and then fix the feature extractor and train the model updater. Scheme 3 jointly trains feature extractor and model updater after separately learning these two components. However, the performance still degrades. The reason is arguably that the ability of the feature extractor for modeling appearance invariance is hampered during joint training since the frame interval between selected images is small.

\subsection{Baseline Comparisons} \label{sec:compare-with-baselines}

\begin{table}[]
	\centering
	\caption{Comparisons with three representative baselines: no update, EMA-based update and SGD-based update. The AUC and EAO metric (higher is better) are reported for OTB and VOT, respectively. For OTB only, the feature extractors are trained with both color and grayscale images. \label{table:baseline-comparison}}
	\resizebox{1\linewidth}{!}{%
		\begin{tabular}{l|ccccccc}
			\hline
								& OTB-2013			& OTB-100			& OTB-50		& VOT-2015		& VOT-2016	& VOT-2017  & Speed(FPS)\\ \hline
			SiamFC-no-update    & 0.608 			& 0.582 			& 0.516 		& 0.290 		& 0.235    	& 0.188 	& 117\\
			SiamFC-ema      	& 0.618 			& 0.597 			& 0.538 			& 0.286 		& 0.259    	& 0.216	& 91	\\
			SiamFC-sgd   		& 0.644 		 & 0.614 			& 0.563 		& 0.306 		& 0.278		& 0.248    	& 13 \\
			SiamFC-lu (Ours)	& \textbf{0.657} & \textbf{0.620} & \textbf{0.577} & \textbf{0.318} & \textbf{0.295} & \textbf{0.263} & 82  \\
			\hline
			CFNet-no-update    	& 0.568 		& 0.541 		& 0.501 		& 0.219 		& 0.201    	& 0.173 & 134	\\
			CFNet-ema      		& 0.608 		& 0.580 		& 0.550 		& 0.237 		& 0.229    	& 0.189	& 79	\\
			CFNet-sgd   		& 0.613 		& 0.590 		& 0.555 		& 0.235 		& 0.212		& 0.182 & 9   	\\
			CFNet-lu (Ours) 	& \textbf{0.621} 	& \textbf{0.599} 	& \textbf{0.565} 	& \textbf{0.242} 		& \textbf{0.230}		& \textbf{0.208}   & 70 	\\ \hline		
		\end{tabular}
	}
\end{table}

We consider three relevant baselines: no update, EMA-based update and SGD-based update. Please refer to Section~\ref{sec:base-tracker-base-update-method} for an introduction of the EMA-based and SGD-based update. 
\begin{itemize}
	\item \textbf{No update}: the target model is initialized in the first frame and then remains fixed.
	\item \textbf{EMA-based update}: the last target model and the current candidate target model are linearly interpolated. The learning rate $\alpha$ is searched on OTB-2013 from 0.01 to 0.2 with step size 0.01.
	\item \textbf{SGD-based update}: we adopt the short-term update and long-term update following MDNet\footnote{Unlike MDNet which updates the last 3 layers, we only update the linear target model, i.e., the last layer for meaningful comparison.} \cite{Nam2016LearningMC}. Short-term update is triggered when the tracker has low confidence while long-term update is conducted every 10 frames. The hyperparameters of the SGD-based update (e.g., online training set size 1000, batch size 8, learning rate 10, number of iterations, 500 for long-term, 200 for short-term, etc.) are searched on OTB-2013.
\end{itemize}

\begin{figure*}
	\centering
	\includegraphics[width=1.3\columnwidth]{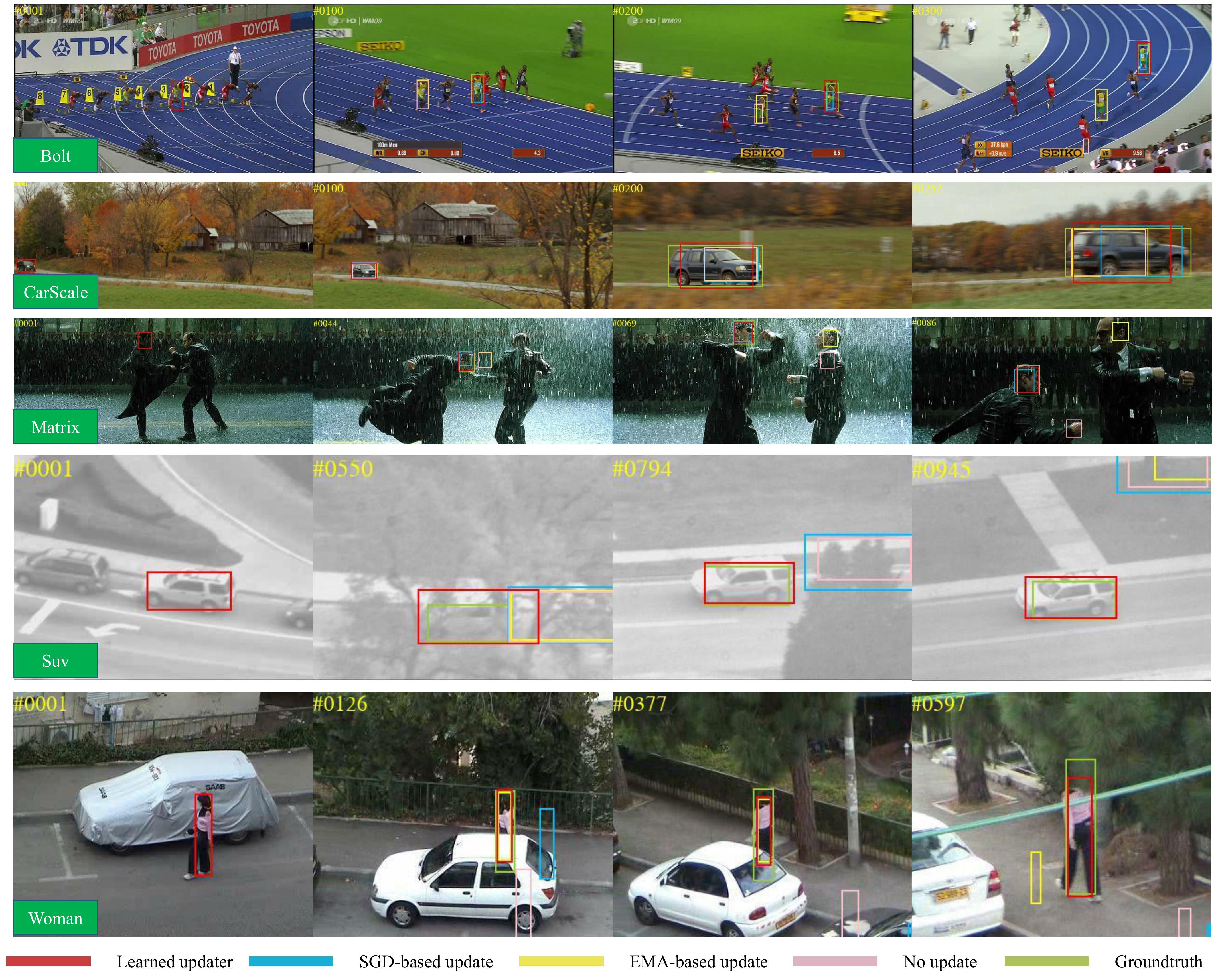}
	\caption{Qualitative results of our learned updater compared with common model update baselines. \textbf{Bolt}: Our learned updater correctly adapts to the target while others are attracted by distractors. SGD-based update method adapts to part of the target instead. \textbf{CarScale}: Other methods keep tracking part of the target since only the front of the car is shown in the first frame. Contrarily, our learned updater gradually adapts to the whole target including both the front and the tail of the car. \textbf{Matrix}: our learned updater is able to perform equally well compared with the SGD-based update in this challenging sequence. \textbf{Suv, Woman}: While all other methods fail, our learned updater still successfully tracks the target.} 
	\label{fig:qualitative-results}
\end{figure*}

For fair comparison, we retrain the updater using publicly available feature extractors (SiamFC and CFNet are open-sourced) with configurations validated in the ablation study. As it turns out, the publicly available feature extractor of SiamFC trained with both color and gray images improves the tracker performance from 0.644 to 0.657 on OTB-2013. This is in line with our observations that the performance of a tracker equipped with our learned updater is positively correlated to that of the feature extractor.
 The results are summarized in Table~\ref{table:baseline-comparison}. Qualitative comparisons are presented in Fig.~\ref{fig:qualitative-results}.

\textbf{Observations}:
1) Our learned updater significantly outperforms the no update and EMA update baselines. 
2) A somewhat surprising yet encouraging result is that our learned updater achieves better performance than the heavily designed and tuned SGD update method. 
3) It is worth noting that for correlation filter-based tracker, EMA update is still a strong baseline.
4) Noticeably, the improvement on SiamFC is consistently larger than that on CFNet. As noted in \cite{valmadre2017end}, the $\mathrm{CF}(\cdot)$ may impose priors that become overly restrictive when enough modeling capacity and data are available.
5) Our learned updater outperforms the SGD-based update method while enjoying comparable efficiency as the EMA-based update method.

\begin{table}[]
	\centering
	\caption{Comparisons with state-of-the-art trackers.}

	\subfloat[\textbf{OTB and VOT:} Trackers are split into two groups: realtime trackers and non-realtime trackers. The AUC and EAO metric (higher is better) are reported for OTB and VOT, respectively. \textcolor{red}{Red} and \textcolor{blue}{blue} fonts indicate \emph{1st} and \emph{2nd} best performance of each group, respectively. \label{table:sota-comparison}]{
		\resizebox{1\linewidth}{!}{%
		\begin{tabular}{l|ccccccc}
			\hline
								& OTB-2013			& OTB-100			& OTB-50		& VOT-2015		& VOT-2016	& VOT-2017  & Speed(FPS)\\ \hline
			ECO 	& \textcolor{red}{0.709}		& \textcolor{red}{0.694}		& \textcolor{blue}{0.643}		& -- 	& \textcolor{red}{0.374}	& \textcolor{blue}{0.280} & 6 \\
			MDNet	& \textcolor{blue}{0.708}		& \textcolor{blue}{0.678}		& \textcolor{red}{0.645}		& \textcolor{red}{0.38}	& --	& -- & 1 \\
			LSART 	& 0.677		& 0.672		& --		& --	& 0.324	& \textcolor{red}{0.323} & 1 \\
			CSRDCF & -- & 0.587 & -- & \textcolor{blue}{0.320} & \textcolor{blue}{0.338} & 0.222 & 13 \\
			CREST 	& 0.673		& 0.623		& -- & -- & 0.283 & -- & 1 \\
			RFL & -- & 0.581 & -- & -- & -- & 0.222 & 15 \\
			\hline
			\hline
			\textbf{SiamFC-lu (Ours)} 	  		& \textcolor{red}{0.657} & \textcolor{blue}{0.620} & \textcolor{red}{0.577} & \textcolor{blue}{0.318} & \textcolor{red}{0.295} & \textcolor{red}{0.263} & 82  \\
			EAST & 0.638 & \textcolor{red}{0.629}  & -- & \textcolor{red}{0.34} & -- & -- & 159 \\
			DSiam   & \textcolor{blue}{0.656} 	& -- 		& -- & 0.293 & -- & -- & 25 \\
			\textbf{CFNet-lu (Ours)} 	  		& 0.621 	& 0.599 	& \textcolor{blue}{0.565} 	& 0.242 		& 0.230		& \textcolor{blue}{0.208}   & 70 	\\
			CFNet    	& 0.610 		& 0.589 		& 0.538 		& -- 		& --    	& -- & 79	\\
			SiamFC    & 0.608 			& 0.582 			& 0.516 		& 0.290 		& \textcolor{blue}{0.235}    	& 0.188 	& 117\\
			\hline
		\end{tabular}
	}
	}

	\subfloat[\textbf{VOT-2017 realtime challenge:} Results of top-performing trackers and our trackers on the VOT-2017 realtime and non-realtime (also called baseline) challenge. The EAO metric (higher is better) is reported. \label{table:vot2017-realtime}]{
		\resizebox{1\linewidth}{!}{%
		\begin{tabular}{l|cccccccc}
			\hline
			challenge				& \textbf{SiamFC-lu (Ours)}	& CSRDCF++	& \textbf{CFNet-lu (Ours)}	& SiamFC	& ECO-HC	& LSART 	& CFCF 		& ECO 	\\ 
			\hline
			realtime    	& \textcolor{red}{0.258} 	& \textcolor{blue}{0.212} 	& 0.200 	& 0.182 	& 0.177    	& 0.055 	& 0.059 	& 0.078 \\
			non-realtime 	& 0.263 	& 0.229 	& 0.208 	& 0.188 	& 0.238		& \textcolor{red}{0.323} 	& \textcolor{blue}{0.286}   	& 0.280	\\
			\hline		
		\end{tabular}}
	}
\end{table}

\subsection{State-of-the-art Comparison}
We compare trackers equipped with our learned updaters against 10 state-of-the-art trackers: ECO\cite{Danelljan2016ECOEC}, MDNet\cite{Nam2016LearningMC}, LSART\cite{sun2018learning}, CSRDCF\cite{lukezic2017discriminative}, CREST\cite{song2017crest}, RFL\cite{yang2017recurrent}, EAST\cite{huang2017learning}, DSiam\cite{guo2017learning}, CFNet\cite{valmadre2017end} and SiamFC\cite{bertinetto2016fully}. The results are summarized in Table~\subref{table:sota-comparison}. 

\textbf{Observations}: 1) SiamFC-lu achieves state-of-the-art performance among realtime trackers. 2) SiamFC-lu outperforms DSiam and RFL, which also focus on improving the update mechanism of SiamFC.

To evaluate the practicability of different tracking methods, VOT-2017 introduces a new ``realtime challenge'' that only allows the tracker to respond in realtime; otherwise, the last predicted target position will be used for the current frame. We compare our method with the state-of-the-art trackers in this setting: CSRDCF++ (a C++ implementation of the CSRDCF\cite{lukezic2017discriminative} tracker), SiamFC\cite{bertinetto2016fully}, ECO-HC (a lightweight version of ECO\cite{Danelljan2016ECOEC} that uses HOG feature), LSART\cite{sun2018learning} and CFCF\cite{gundogdu2018good}. The results are shown in Table~\subref{table:vot2017-realtime}. It can be observed that our approach achieves state-of-the-art results in the realtime setting.

\section{Discussion}
One interesting question would be why learning to update actually works? We try to answer this question in three different perspectives. 1) From the high-level perspective, fast model update is viable because videos are intrinsically structured (e.g., temporal dependencies between target features, target variation patterns). Our learned updater captures these structures in a data driven manner. 2) As for the functionality, our learned updater can be seen as a learnable extension of the EMA-based update. The difference is that, instead of linearly interpolating the last target model and the current candidate target model, we adopt the gating mechanism.  As a result, the learned updater inherits the efficiency of EMA-based update and the effectiveness of learning based method. 3) Empirically speaking, as shown in Fig.~\ref{fig:qualitative-results}, after being trained under the classification loss and anchor loss, our learned updater is able to reliably absorb target variations while resisting the distractors. 

As a first attempt, however, there are still many interesting problems left uninvestigated. One issue of RNN-based updater is that, the convolutional GRU requires large amount of GPU memory during offline training and thus being difficult to apply to target models with large numbers of parameters. How to scale to large target models would be an interesting research problem. Moreover, in this work, only linear target models are considered, how to extend to the non-linear cases such as MDNet is valuable future work. 

\section{Conclusion}
We propose a learning based framework to tackle the problem of model update during tracking. As a first attempt, only the update of linear target model is considered. The learned updater is parameterized based on RNN and successfully learned with several techniques proposed in this work. Our learned updater outperforms two common model update baselines including the efficient EMA-based update and the well-designed SGD-based update. After offline training, our learned updater can run efficiently during testing; therefore, our learned updater is able to consistently improve the base trackers without sacrificing the speed. Notably, the SiamFC tracker has been improved by nearly 40\% in terms of the EAO on VOT-2017 while running at the speed of 82fps, which achieves state-of-the-art performance among realtime counterparts. In the future, we plan to extend the learning to update paradigm to non-linear target models.

\appendices

\section*{Acknowledgment}
This work was supported in part by National Natural Science Foundation of China (grant No. 61733007, 61572207). 

The authors would like to thank Chong Luo and Anfeng He for helpful discussions.

\ifCLASSOPTIONcaptionsoff
  \newpage
\fi

\bibliographystyle{IEEEtran}
\bibliography{learn_to_update}

\begin{IEEEbiography}[{\includegraphics[width=1in,height=1.25in,clip,keepaspectratio]{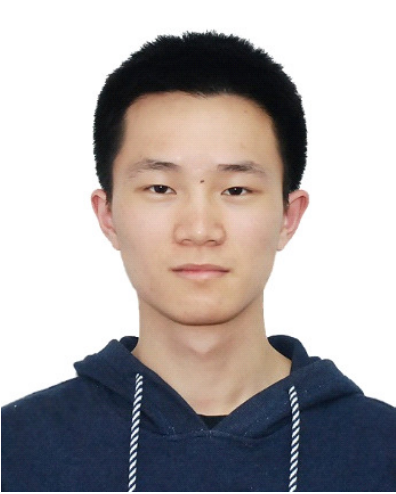}}]{Bi Li}
received the B.Sc. degree from Huazhong University of Science and Technology (HUST), Wuhan, China, in 2014, and is currently a Ph.D. student at the media and communication lab, HUST, supervised by Prof. Wenyu Liu. His research interests include meta-learning, few-shot learning and object tracking.
\end{IEEEbiography}

\begin{IEEEbiography}[{\includegraphics[width=1in,height=1.25in,clip,keepaspectratio]{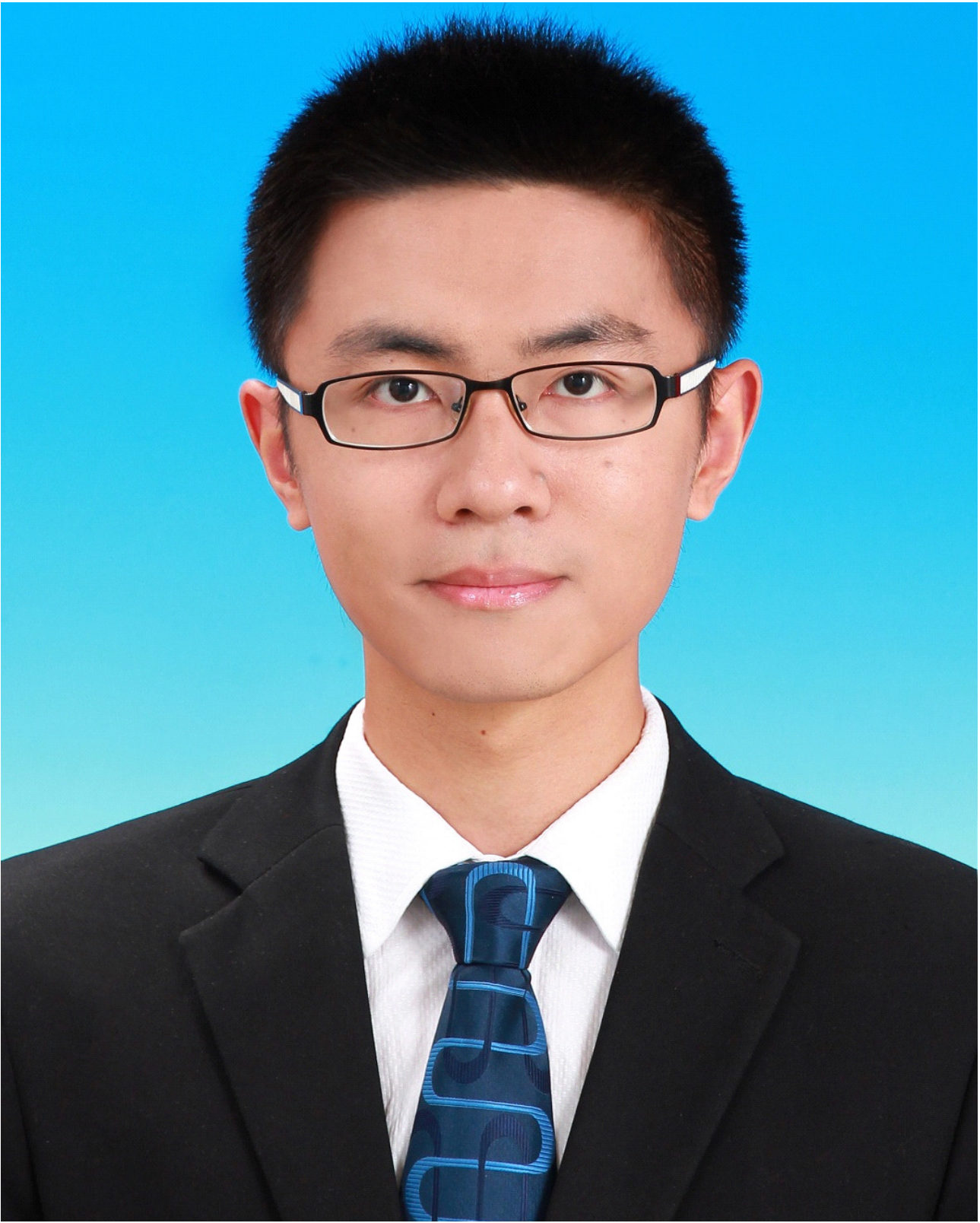}}]{Wenxuan Xie}
	received the B.Sc. degree from Nanjing University, Nanjing, China, in 2010, and the Ph.D. degree from Peking University, Beijing, China, in 2015. He has been working as an associate researcher in Microsoft Research Asia since 2015. His research interests include computer vision and machine learning.
\end{IEEEbiography}

\begin{IEEEbiography}[{\includegraphics[width=1in,height=1.25in,clip,keepaspectratio]{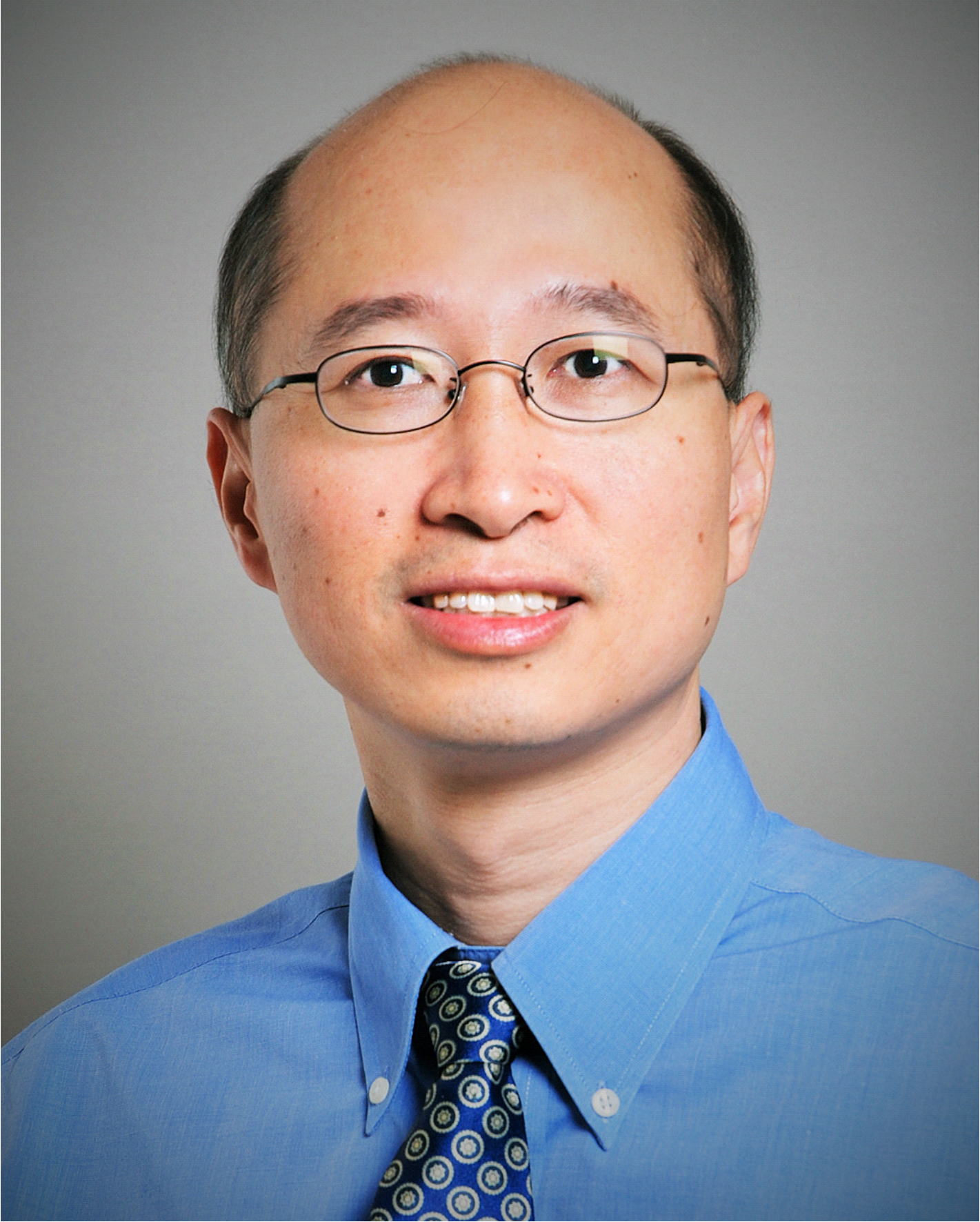}}]{Wenjun (Kevin) Zeng}
	 (M'97-SM'03-F'12) is a Principal Research Manager and a member of the senior leadership team (SLT) of Microsoft Research Asia. He has been leading the video analytics research empowering the Microsoft Cognitive Services and Azure Media Analytics Services since 2014. He was with Univ. of Missouri (MU) from 2003 to 2016, most recently as a Full Professor. Prior to that, he had worked for PacketVideo Corp., Sharp Labs of America, Bell Labs, and Panasonic Technology. Wenjun has contributed significantly to the development of international standards (ISO MPEG, JPEG2000, and OMA).  He received his B.E., M.S., and Ph.D. degrees from Tsinghua Univ., the Univ. of Notre Dame, and Princeton Univ., respectively.  His current research interest includes mobile-cloud media computing, computer vision, social network/media analysis, and multimedia communications and security.
	
	He was an Associate Editor-in-Chief of IEEE Multimedia Magazine, and was an AE of IEEE Trans. on Circuits \& Systems for Video Technology (TCSVT), IEEE Trans. on Info. Forensics \& Security, and IEEE Trans. on Multimedia (TMM). He was a Special Issue Guest Editor for the Proceedings of the IEEE, TMM, ACM TOMCCAP, TCSVT, and IEEE Communications Magazine. He was on the Steering Committee of IEEE Trans. on Mobile Computing and IEEE TMM. He served as the Steering Committee Chair of IEEE ICME in 2010 and 2011, and is serving or has served as the General Chair or TPC Chair for several IEEE conferences (e.g., ICME'2018, ICIP'2017).  He was the recipient of several best paper awards. He is a Fellow of the IEEE.
\end{IEEEbiography}

\begin{IEEEbiography}[{\includegraphics[width=1in,height=1.25in,clip,keepaspectratio]{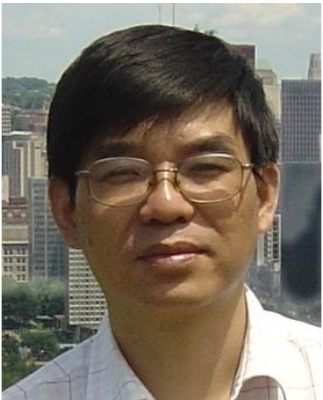}}]{Wenyu Liu}
	(M'08-SM'15) received the B.S. degree in Computer Science from Tsinghua University, Beijing, China, in 1986, and the M.S. and Ph.D. degrees, both in Electronics and Information Engineering, from Huazhong University of Science \& Technology (HUST), Wuhan, China, in 1991 and 2001, respectively. He is now a professor and associate dean of the School of Electronic Information and Communications, HUST. His current research areas include computer vision, multimedia, and machine learning. He is a senior member of IEEE.
\end{IEEEbiography}

\end{document}